\documentclass{article} 
\usepackage{iclr2027_conference,times}

\usepackage{amsmath,amsfonts,bm}

\def\eqref#1{equation~\ref{#1}}

\def\1{\bm{1}}

\DeclareMathAlphabet{\mathsfit}{\encodingdefault}{\sfdefault}{m}{sl}
\SetMathAlphabet{\mathsfit}{bold}{\encodingdefault}{\sfdefault}{bx}{n}

\usepackage{graphicx}
\usepackage{booktabs}
\usepackage{multirow}
\usepackage{subcaption}
\usepackage{xcolor}
\usepackage{colortbl}
\usepackage{xspace}
\usepackage{algorithm}
\usepackage{algpseudocode}
\usepackage{listings}

\usepackage{hyperref}
\usepackage{url}
\usepackage{nicefrac}

\title{LabBook: Harnessing Experimental History for Efficient
LLM-Driven Discovery}

\author{%
\parbox{\textwidth}{\centering
Bo Yuan\textsuperscript{1} \quad
Wenqian Ye\textsuperscript{2} \quad
Zelin Zhao\textsuperscript{1} \quad
Lama Moukheiber\textsuperscript{1} \\
Henry Kautz\textsuperscript{2} \quad
Aidong Zhang\textsuperscript{2} \quad
Yongxin Chen\textsuperscript{1} \\[0.4em]
{\normalfont\textsuperscript{1}Georgia Institute of Technology
} \\
{\normalfont\textsuperscript{2}University of Virginia, Charlottesville
}
}
}

\iclrfinalcopy

\begin{document}

\maketitle
\lhead{} 

\begin{abstract}

Evolutionary approaches to LLM-driven discovery often generate new programs
from a small set of selected ancestors. This keeps contexts manageable but can
omit useful evidence from other experiments, whereas including the full
experimental history produces long, redundant contexts. We introduce a
simple, single-agent discovery harness built around \textbf{LabBook}, an
agent-maintained memory that serves two complementary roles: guiding retrieval of relevant
evidence from a complete experimental log and informing the generation of new
solutions. At each iteration, the same agent combines its memory with retrieved
evidence and jointly produces the next program and an updated LabBook. This separates complete history retention from
selective context construction, without requiring an explicit population or
branching search structure. On 49 Frontier-CS problems, LabBook improves the observed quality–cost trade-off over the evaluated evolutionary baselines with two backbones, while remaining competitive across nine additional mathematical, systems, and heuristic-design tasks.

Code will be released at \url{https://github.com/BoYuanVisionary/LabBook}.
\end{abstract}

\section{Introduction}
\label{sec:introduction}

Large language models are increasingly used as engines for scientific
discovery. In program-based discovery, an LLM proposes an executable
hypothesis, such as an algorithm, heuristic, or systems design; an evaluator
measures its behavior; and a discovery harness uses the result to decide what
the model should try next. This feedback-driven paradigm
has produced new mathematical constructions, competitive heuristics, and
improvements to systems code
\citep{romeraparedes2024funsearch,liu2024eoh,ye2024reevo,
novikov2025alphaevolve,lange2026shinkaevolve}. As model capabilities continue
to improve, recent systems have extended this loop to the improvement of
coding agents and exploration policies themselves
\citep{zhang2026dgm,zheng2026dreamrsi}. The harness---and particularly how it
carries information across experiments---therefore becomes an increasingly
important determinant of scientific discovery quality and cost.

Most existing harnesses organize discovery as an explicit evolutionary search.
They maintain a population or archive of evaluated programs, select one or a
few promising ancestors, and ask the LLM to mutate or recombine them
\citep{novikov2025alphaevolve,lange2026shinkaevolve,
assumpcao2026codeevolve,cemri2026adaevolve,liu2026evox}. This structure keeps
each generation context manageable and preserves working code along successful
lineages. However, selection is also a lossy form of experimental memory: an
attempt that is not chosen as a parent can disappear from the next proposal
context even when it reveals a useful failure mode, a partially successful
idea, or an implementation detail needed to interpret later results.

Providing the complete experimental history is the opposite extreme. It avoids
discarding evidence, but its context grows with every attempt and repeatedly
includes large programs, similar failures, and obsolete hypotheses. Prior work
shows both that evaluated solutions can guide in-context optimization
\citep{yang2024opro} and that long contexts do not guarantee effective use of
the relevant information \citep{liu2024lostmiddle}. The resulting design
problem is therefore not simply whether to remember more or less. A discovery
harness should retain the complete experimental record while constructing a
small, targeted context for each new proposal---much as a researcher keeps a
notebook but consults only the evidence relevant to the current decision.

\begin{figure*}[t]
    \centering
    \includegraphics[width=0.96\textwidth]{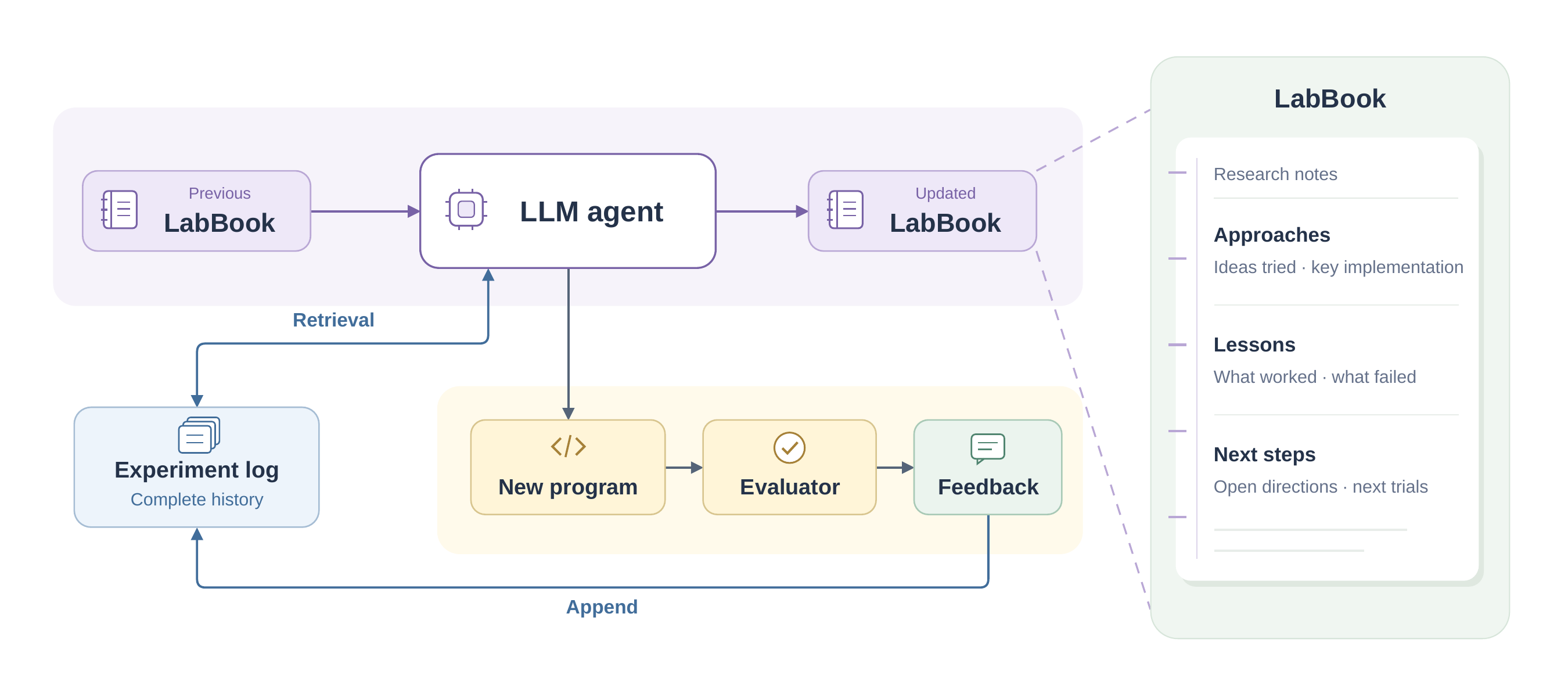}
    \caption{Overview of the LabBook harness. At each iteration, a single agent
    combines the task, the latest evaluated attempt, and the compact LabBook
    state; it may retrieve exact evidence from the append-only experimental
    log before jointly producing the next program and a rewritten LabBook.
    Evaluation appends a new record to the log, closing the discovery loop.}
    \label{fig:labbook-overview}
\end{figure*}

We introduce \textbf{LabBook}, an agent-maintained memory that separates
history retention from context construction. The harness records every
evaluated proposal and its feedback in a lossless experimental log.
The agent then uses LabBook to retrieve relevant records and
synthesize a focused context for generating the next solution. The evaluation
of that solution adds new evidence to the log, and the agent updates
LabBook, creating an online loop in which discovery experience
recursively improves future discovery. Because information from earlier
experiments is passed forward through LabBook rather than through
selected parent programs, the harness requires no explicit population or
branching search structure. Architecturally, the harness consists of one agent, one rewritten natural-language state, and a fixed read-only interface to an append-only log. It avoids search-specific sampler design and hyperparameters, such as population size, parent-selection rules, and branching schedules as well as multi-agent pipelines with separate planner, reflector, and generator roles. Figure~\ref{fig:labbook-overview} illustrates this interaction.

We evaluate the LabBook on 49 open-ended
Frontier-CS problems~\citep{mang2025frontiercs} and 9 additional mathematical and heuristic-design tasks. Across
these settings, it achieves comparable or better solution quality than the
evaluated baselines at lower measured cost. Our contributions
are:

\begin{itemize}
    \item We identify the coupling of experimental retention and proposal-context
    construction as a central information bottleneck in LLM-driven scientific
    discovery.
    \item We develop a simple harness built around LabBook, which combines
    complete experimental logging with selective retrieval and compact context
    synthesis, without an explicit population or branching search structure.
    \item We evaluate the resulting quality--cost trade-off across 58 discovery
    problems and analyze how accumulated evidence guides subsequent proposals.
\end{itemize}

\section{Related Work}
\label{sec:related_work}

\subsection{Evolutionary and Self-Improving Scientific Discovery}

Program-based LLM scientific discovery commonly embeds a language model in an
evolutionary loop. FunSearch combines program generation, programmatic
evaluation, and an island-based database that supplies high-scoring programs
to subsequent generations \citep{romeraparedes2024funsearch}. EoH jointly
represents each heuristic as natural-language thought and executable code
\citep{liu2024eoh}, while ReEvo adds short- and long-term reflections as verbal
gradients for population-based heuristic evolution \citep{ye2024reevo}.
AlphaEvolve scales program archives, evaluator feedback, and direct code
modification to broad mathematical and systems problems
\citep{novikov2025alphaevolve}. ShinkaEvolve, CodeEvolve, AdaEvolve, and EvoX
further improve parent selection, novelty, quality-diversity, resource
allocation, or adaptation of the search strategy
\citep{lange2026shinkaevolve,assumpcao2026codeevolve,
cemri2026adaevolve,liu2026evox}. These methods differ substantially in their
search policies, but each proposal is constructed from a selected view of the
program archive or lineage.

A related line makes the discovery process itself an object of improvement.
The Darwin G\"odel Machine evolves the code of coding agents while retaining an
open-ended archive of agent variants \citep{zhang2026dgm}. DeltaEvolve replaces
full-code history with structured semantic deltas intended to preserve the
causes of performance changes \citep{jiang2026deltaevolve}. Dream-RSI treats
completed discovery trees as replay simulators, using inexpensive off-policy
feedback to improve an explicit exploration policy before redeploying it
online \citep{zheng2026dreamrsi}. Our harness instead leaves the generator and
its parameters fixed and uses LabBook to manage the within-run
information path: a complete experiment log is retained, while the memory
agent decides which evidence should enter the next proposal context.

\subsection{Memory and Experimental-History Reuse}

Language-agent research has developed several ways to learn from interaction
history without updating model parameters. Generative Agents combine a
natural-language memory stream with reflection and dynamic retrieval
\citep{park2023generativeagents}; Reflexion stores linguistic feedback for
subsequent trials \citep{shinn2023reflexion}; and ExpeL extracts reusable
insights from collections of trajectories \citep{zhao2024expel}. ReasoningBank
distills strategies from both successful and failed experience and retrieves
them on later tasks \citep{ouyang2026reasoningbank}. MemGPT frames limited
context as hierarchical memory management \citep{packer2024memgpt}, while
long-context studies show that simply exposing a model to more text does not
guarantee effective use of the relevant evidence \citep{liu2024lostmiddle}.
OPRO provides the complementary optimization extreme by appending prior
solutions and their scores directly to the prompt \citep{yang2024opro}.
Among memory-aware discovery systems,
CausalEvolve combines a causal scratchpad with explicit parent and inspiration
selection \citep{chen2026causalevolve}, while RefineEvo retains
population-based evolution and introduces separate Planner, Evolver, and
Reflector roles together with a bidirectional experience pool
\citep{wu2026refineevo}. In contrast, our lightweight harness consists of one
agent, one rewritten LabBook state, and four read-only operations over an
append-only lossless log. It requires no population or branching management,
no search-specific sampler hyperparameters, and no separate planner,
reflector, and generator roles or learned retriever. The same agent chooses
which evidence to inspect and jointly produces the next program and updated
LabBook.

\section{LightWeight Harness Design}
\label{sec:method}

\subsection{Discovery as a History-to-Proposal Mapping}

We consider a discovery task defined by a description $\tau$, a program space
$\mathcal{X}$, and an evaluator $\mathcal{E}$. At iteration $t$, the agent
produces a complete program $x_t\in\mathcal{X}$, and the evaluator returns
$y_t=\mathcal{E}(x_t)=(s_t,\phi_t),$
where $s_t$ is the objective score and $\phi_t$ contains auxiliary metrics and
textual feedback. After $T$ iterations, the harness selects
$ t^\star=\arg\max_{1\leq t\leq T}s_t,  x^\star=x_{t^\star}.$
At iteration $t$, all previously evaluated solutions and feedback form the
ordered history $\mathcal{H}_{t-1}=[(x_i,y_i)]_{i=1}^{t-1}$. We write a generic
LLM-driven discovery harness as
\begin{equation}
    c_t=\Gamma(\mathcal{H}_{t-1};\tau),
    \qquad x_t\sim A_\theta(\,\cdot\mid\tau,c_t),
    \label{eq:history-proposal-map}
\end{equation}
where $\Gamma$ maps experimental history to the context used for the next
proposal and $A_\theta$ stands for the agent. This decomposition separates two
decisions that are often coupled: how past experiments are represented and
exposed, and how the agent converts the resulting context into a new program.
We therefore treat the history-to-context mapping $\Gamma$ as a first-class design object for
LLM-driven discovery.

Most population-based methods first select an archive subset, sample one or more
parents, and place their programs and scores in the generation context
\citep{romeraparedes2024funsearch,novikov2025alphaevolve,
lange2026shinkaevolve}. A full-history method instead sets
$\Gamma_{\mathrm{full}}(\mathcal{H})=\mathcal{H}$. Our harness realizes $\Gamma$ through LabBook: one LLM agent maintains a compact state, selectively retrieves exact evidence from the complete history,
and jointly produces the next program and updated memory. No explicit
population or branching structure is used to construct $c_t$.

\subsection{LabBook}
\label{section:LabBook}

\paragraph{Intuition.}
LabBook mirrors how a researcher manages a long sequence of
experiments. A researcher neither keeps every implementation detail in working
memory nor forgets an unsuccessful direction once it is abandoned. Instead,
they continually consolidate what has been tried, what worked or failed, and
which directions remain promising, while returning to the experimental record
when an exact implementation or failure case becomes relevant. LabBook
plays this intermediate role between the accumulated experimental history and
the next proposal. Viewed through the lens of memory systems, the lossless experimental log provides an episodic record of individual trials, while LabBook consolidates their higher-level lessons into a compact semantic state.

Formally, in our implementation, LabBook is the agent-maintained state that carries discovery experience
between iterations. It is a single free-form natural-language document rather
than a list of previous programs or a fixed-schema database. The agent is
instructed to maintain an approach-level account containing: (i) the idea
behind each attempted approach, (ii) decisive
implementation details and parameters, and (iii) the lesson learned and the
most useful unresolved direction. LabBook serves as the key component to construct the history-to-proposal mapping $\Gamma$. Our implementation realizes $\Gamma$ in two stages:
\emph{memory-guided context initialization} and
\emph{agent-guided evidence retrieval}.

\paragraph{Memory-guided context initialization.}
Before retrieval, the harness constructs the history-dependent part of the
context as
\begin{equation*}
    c_t^{(0)}
    =\left((x_{t-1},y_{t-1}),m_{t-1},d_t\right),
    \label{eq:initial-context}
\end{equation*}
which is supplied to the agent together with the task description $\tau$ from
Equation~\ref{eq:history-proposal-map}. The first component is the latest
evaluated experiment. Its complete program $x_{t-1}$ and evaluation
$y_{t-1}$ are always included, allowing local refinement without a retrieval
call. The second component is LabBook $m_{t-1}$, which carries the
agent's persistent semantic account of earlier experiments. The final
component contains deterministic progress signals
\begin{equation*}
        s^\star_{t-1}=\max_{i<t}s_i, \quad i^\star_{t-1}=\min\{i<t:s_i=s^\star_{t-1}\},
\end{equation*}
where $d_t =(s^\star_{t-1},i^\star_{t-1})$ represents the effort since the current best was first
attained. These signals expose regressions and plateaus, but do not require the agent to modify the best program. We assume that higher scores indicate better solutions; for minimization problems, we replace \(\max\) with \(\min\).

\paragraph{Agent-guided evidence retrieval.}
Starting from $c_t^{(0)}$, the agent decides whether additional historical
evidence is needed and, if so, which evidence to inspect. Retrieval uses a
coarse-to-fine, read-only interface rather than embeddings or a learned
retriever. Its action space is
\begin{equation}
    q_{t,j}\in
    \left\{
    \textsc{History},\,
    \textsc{Code}(i),\,
    \textsc{Result}(i),\,
    \textsc{Done}
    \right\}.
    \label{eq:retrieval-actions}
\end{equation}
\textsc{History} returns a lightweight iteration-indexed overview
$[(i,s_i)]_{i<t}$ containing scores. Given an
iteration number, \textsc{Code}$(i)$ returns the complete program $x_i$, while
\textsc{Result}$(i)$ returns its full evaluator feedback
$y_i$. \textsc{Done} terminates retrieval. At retrieval step $j$, the agent chooses a query conditioned on all observations
returned so far. The agent may stop without retrieval when its initial context is sufficient,
and otherwise issues at most $K$ queries. Large programs and detailed
evaluator outputs therefore enter the active context only when selected by the
agent. After $J_t$ queries, the context used for generation is
$c_t=c_t^{(J_t)}$, and the agent emits both a complete program
and an update for LabBook:
\begin{equation}
    (x_t,m_t)\sim
    A_\theta(\,\cdot\mid\tau,c_t^{(J_t)}).
    \label{eq:joint-update}
\end{equation}
The program is then evaluated, and its record is appended
to the experimental log $\mathcal{H}_{t-1}$.

\paragraph{Why simple retrieval is sufficient.}
LabBook provides a high-level semantic map of the search: it records which approaches have been explored, what was learned from them, and which directions remain promising. Moreover, retrieval takes place within a
single discovery run, where the task specification, program interface, and
evaluator remain fixed. The resulting log is therefore a homogeneous,
chronologically indexed collection of experiments rather than an open-domain
document corpus.

\subsection{Harness Design}

Algorithm~\ref{alg:discovery-memory} combines the two stages above into the
complete harness. Each iteration constructs the initial context described in
Section~\ref{section:LabBook}, lets the agent augment it through a bounded number of read-only
queries, and evaluates exactly one newly generated program. The retrieval
budget is an upper bound rather than a required number of calls:
\textsc{Done} terminates retrieval as soon as the current context is
sufficient. In our implementation, the complete history $\mathcal{H}_{t-1}$ is stored locally, while LabBook is rewritten at each iteration, allowing the active context to remain compact.

\begingroup
\algrenewcommand{\algorithmiccomment}[1]{\hspace{0.75em}$\triangleright$ #1}
\begin{algorithm}[t]
\caption{LabBook harness}
\label{alg:discovery-memory}
\begin{algorithmic}[1]
\Require Task $\tau$, evaluator $\mathcal{E}$, agent $A_\theta$, budget $T$,
tool limit $K$
\State $m_0\gets\emptyset$; $\mathcal{H}_0\gets[\,]$
    \Comment{Initialize compact state and exact log}
\For{$t=1,\ldots,T$}
    \State $d_t\gets(s^\star_{t-1},i^\star_{t-1})$
        \Comment{Expose best-so-far progress}
    \State Construct $c_t^{(0)}$ from the latest experiment,
    $m_{t-1}$, and $d_t$
    \For{$j=1,\ldots,K$}
        \State $q_{t,j}\sim A_\theta(\,\cdot\mid\tau,c_t^{(j-1)})$
            \Comment{Agent selects the next read}
        \If{$q_{t,j}=\textsc{Done}$}
            \State \textbf{break}
        \EndIf
        \State $o_{t,j}\gets
        \mathcal{R}(\mathcal{H}_{t-1},q_{t,j})$
            \Comment{Read exact historical evidence}
        \State $c_t^{(j)}\gets
        c_t^{(j-1)}\mathbin{\Vert}(q_{t,j},o_{t,j})$
    \EndFor
    \State $(x_t,m_t)\sim
    A_\theta(\,\cdot\mid\tau,c_t^{(J_t)})$
        \Comment{Joint proposal and memory rewrite}
    \State $y_t\gets\mathcal{E}(x_t)$
    \State $\mathcal{H}_t\gets
    \mathcal{H}_{t-1}\mathbin{\Vert}(x_t,y_t)$
\EndFor
\State $t^\star\gets\arg\max_{1\leq t\leq T}s_t$
\State \Return $x_{t^\star}$ \Comment{Return the best evaluated program}
\end{algorithmic}
\end{algorithm}
\endgroup

\section{Experiments}
\label{sec:experiments}

\subsection{Experimental Setup}

\paragraph{Tasks.}
We select tasks from four benchmarks. {Frontier-CS} is an open-ended
benchmark of verifiable computer-science problems with continuous
partial-credit evaluation \citep{mang2025frontiercs}. From its algorithmic
track, we randomly sampled 49 problems. We additionally retain nine tasks from three public sources.
From the mathematical problems studied by AlphaEvolve
\citep{novikov2025alphaevolve}, we use \texttt{circle\_packing} and
\texttt{heilbronn\_convex}. From the AI-Driven Research for Systems suite
\citep{cheng2025barbarians}, we use \texttt{eplb}, \texttt{prism}, and
\texttt{txn\_scheduling}. From HeuriGym \citep{chen2026heurigym}, we use
\texttt{egraph\_extraction}, \texttt{operator\_scheduling},
\texttt{pedigree}, and \texttt{pickup\_delivery\_time\_windows}. These tasks
span mathematical construction, systems optimization, compilers, electronic
design automation, computational biology, and vehicle routing. See Appendix~\ref{app:benchmark-composition} for detailed descriptions of selected problems.

\paragraph{Baselines.}
We compare against three evolutionary methods: OpenEvolve \citep{openevolve2025}, AdaEvolve
\citep{cemri2026adaevolve}, and EvoX \citep{liu2026evox}. All methods
use the same underlying LLM and evaluator. We use their open-source implementation from SkyDiscover~\citep{liu2026skydiscover}.
All baselines use the default hyperparameters provided by SkyDiscover.

\paragraph{Implementation.}
We evaluate every method with two backbone language models: DeepSeek V4 Flash
\citep{xu2026deepseek} and Qwen 3.6 Flash
\citep{qwen2026qwen36flash}. We use a temperature of $0.7$ for all methods. Within each task--model pair, all
methods receive the same task prompt, initial program and evaluator. The maximum output length is 32k tokens and evaluator timeouts follow the original benchmarks.
LabBook allows at most $K=5$ read-only retrieval actions per
iteration. The implementation supports a configurable character-level
truncation for memory, but we observe no truncation; empirical memory lengths are
reported in Appendix~\ref{app:memory-size}.

\subsection{Evaluation Metrics}

\paragraph{Solution quality.}
We follow the evaluation protocol released with each benchmark. Frontier-CS
assigns every evaluated program a normalized partial-credit score
$s_p(x)\in[0,100]$ \citep{mang2025frontiercs}. We report the
mean and median best-so-far score across its 49 problems. For the two AlphaEvolve problems, we retain their geometric objectives. For the other problesm from ADRS \citep{cheng2025barbarians} and HeuriGym \citep{chen2026heurigym}, we use their released task-specific scores.

\paragraph{Discovery cost.}
Prior work commonly uses either iteration count or token count as the discovery
budget \citep{novikov2025alphaevolve,cemri2026adaevolve,liu2026evox}. However, neither
is directly comparable across harnesses: an iteration may invoke different
numbers of model calls and consume different amounts of context, while input,
cached-input, and output tokens have different prices. We therefore use
measured API cost as our primary efficiency metric. See Appendix~\ref{app:additional-metrics} for rates. Our main results therefore report best-so-far quality as a
function of cumulative dollar cost. Token counts are reported in
Appendix~\ref{app:additional-metrics}.

\subsection{Main Results}

\begin{figure*}[t]
    \centering
    \includegraphics[width=\textwidth]{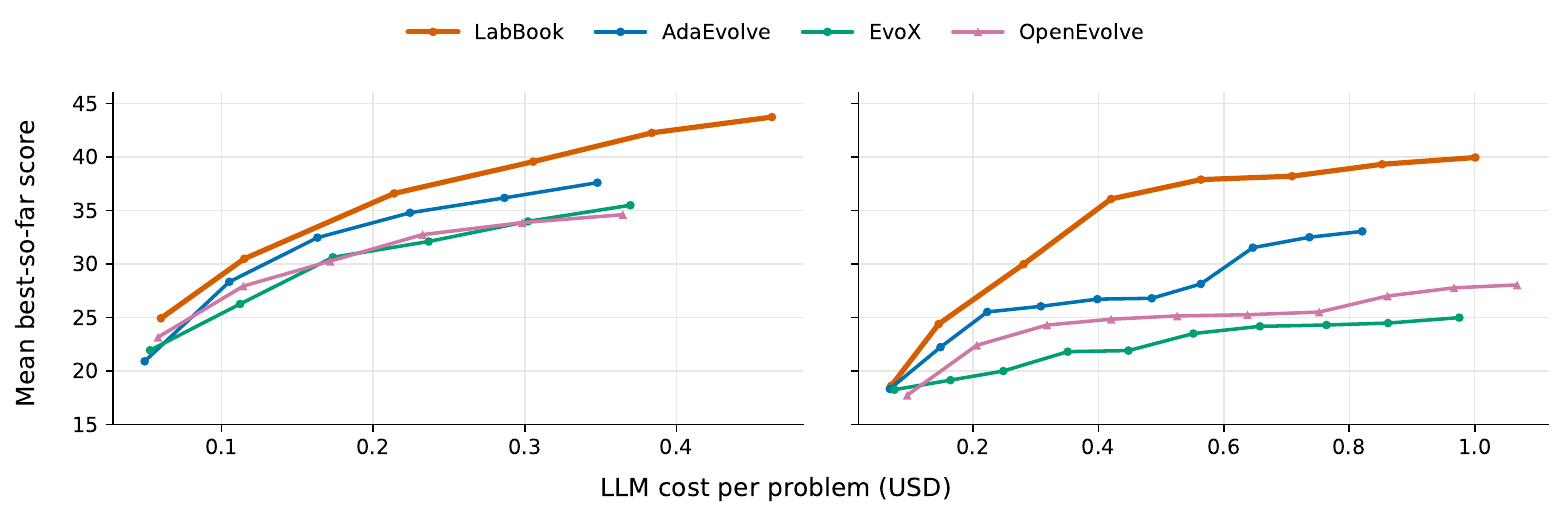}
    \caption{Cost--performance on Frontier-CS with DeepSeek V4 Flash (left)
    and Qwen 3.6 Flash (right), averaged over three independent runs. Each
    point averages the best-so-far score and cumulative cost across the 49
    problems. LabBook is shown after 5 iterations and every 10
    iterations thereafter; the baselines are shown every 10 iterations.}
    \label{fig:frontier-cost-performance}
\end{figure*}

\paragraph{Frontier-CS.}
We compare LabBook with the baselines on the same 49 retained
Frontier-CS problems at matched cost.
As shown in Figure~\ref{fig:frontier-cost-performance}, LabBook achieves
the strongest cost--performance trade-off with both backbone models and
maintains a higher mean best-so-far score over most of the measured budget.
The corresponding median scores exhibit the same overall ordering
(Appendix~\ref{app:frontier-median}).
At the final displayed budgets, LabBook improves over the strongest
baseline by 6.13 points (16.3\%) with DeepSeek and 6.91 points (20.9\%) with
Qwen. Using the underlying per-iteration trajectories, it surpasses the final
AdaEvolve score at a cost of \$0.243 rather than \$0.348 per problem with
DeepSeek, and \$0.420 rather than \$0.821 with Qwen, corresponding to 30.2\%
and 48.8\% lower cost, respectively.


Table~\ref{tab:cross-domain-results} reports the best task-native score and
total measured cost for each run. LabBook obtains the best or tied-best
score on all nine tasks with DeepSeek V4 Flash and on seven of nine tasks with
Qwen 3.6 Flash.

\begin{table*}[t]
\centering
\caption{Comprehensive evaluation on nine cross-domain discovery tasks. Score
reports the task-native objective, and cost is the cumulative measured API cost
in USD at the reported checkpoint. Higher scores are better for the first six
tasks. For the final three tasks, E-Graph Extraction reports total extraction
cost, Operator Scheduling reports total latency, and Pedigree reports total
genotype changes; lower is better for all three. Bold denotes the best score
for each task and backbone, and underlining denotes the second-best distinct score. All methods use 50 iterations. 
}
\label{tab:cross-domain-results}
\setlength{\tabcolsep}{3.5pt}
\renewcommand{\arraystretch}{0.9}
\begin{tabular}{p{0.25\textwidth}lllll}
\toprule
\multirow{2}{*}{Task} & \multirow{2}{*}{Method} &
\multicolumn{2}{c}{DeepSeek V4 Flash} &
\multicolumn{2}{c}{Qwen 3.6 Flash} \\
\cmidrule(lr){3-4}\cmidrule(lr){5-6}
& & Score & Cost $\downarrow$ & Score & Cost $\downarrow$ \\
\midrule
\multirow{4}{0.25\textwidth}{Circle Packing}
& OpenEvolve    & 0.9288 & 0.283 & 0.8880 & 0.395 \\
& EvoX          & 0.8583 & 0.229 & 0.7942 & 0.301 \\
& AdaEvolve     & \underline{0.9934} & 0.562 & \textbf{0.9510} & 0.403 \\
\rowcolor{yellow!12}
\cellcolor{white}& \textbf{LabBook (Ours)} & \textbf{1.0004} & 0.207 & \underline{0.9347} & 0.816 \\
\midrule
\multirow{4}{0.25\textwidth}{Heilbronn Convex}
& OpenEvolve    & \underline{0.7283} & 0.205 & 0.7519 & 0.548 \\
& EvoX          & 0.6305 & 0.246 & 0.5122 & 0.357 \\
& AdaEvolve     & 0.7167 & 0.480 & \textbf{0.7758} & 0.434 \\
\rowcolor{yellow!12}
\cellcolor{white}& \textbf{LabBook (Ours)} & \textbf{0.9452} & 0.179 & \underline{0.7740} & 0.475 \\
\midrule
\multirow{4}{0.25\textwidth}{EPLB}
& OpenEvolve    & 0.1289 & 0.300 & \underline{0.1287} & 0.476 \\
& EvoX          & 0.1288 & 0.317 & 0.1275 & 0.303 \\
& AdaEvolve     & \underline{0.1291} & 0.432 & 0.1282 & 0.735 \\
\rowcolor{yellow!12}
\cellcolor{white}& \textbf{LabBook (Ours)} & \textbf{0.1449} & 0.299 & \textbf{0.1447} & 0.664 \\
\midrule
\multirow{4}{0.25\textwidth}{PRISM}
& OpenEvolve    & 24.1089 & 0.119 & \underline{25.6236} & 0.188 \\
& EvoX          & \underline{26.1214} & 0.152 & 22.6817 & 0.139 \\
& AdaEvolve     & \textbf{26.2560} & 0.271 & \textbf{26.2560} & 0.152 \\
\rowcolor{yellow!12}
\cellcolor{white}& \textbf{LabBook (Ours)} & \textbf{26.2560} & 0.197 & \textbf{26.2560} & 0.511 \\
\midrule
\multirow{4}{0.25\textwidth}{Transaction Scheduling}
& OpenEvolve    & \underline{3802.28} & 0.477 & 3731.34 & 0.436 \\
& EvoX          & 3267.97 & 0.225 & 3546.10 & 0.281 \\
& AdaEvolve     & 3787.88 & 0.484 & \underline{3816.79} & 0.425 \\
\rowcolor{yellow!12}
\cellcolor{white}& \textbf{LabBook (Ours)} & \textbf{3861.00} & 0.288 & \textbf{4166.67} & 0.689 \\
\midrule
\multirow{4}{0.25\textwidth}{Pickup--Delivery with Time Windows}
& OpenEvolve    & 0.5000 & 0.260 & \underline{0.5000} & 0.573 \\
& EvoX          & 0.6589 & 0.397 & \underline{0.5000} & 0.493 \\
& AdaEvolve     & \underline{0.7805} & 0.422 & \underline{0.5000} & 0.680 \\
\rowcolor{yellow!12}
\cellcolor{white}& \textbf{LabBook (Ours)} & \textbf{0.9964} & 0.356 & \textbf{0.6613} & 0.788 \\
\midrule\midrule
\multirow{4}{0.25\textwidth}{E-Graph Extraction}
& OpenEvolve    & \underline{19642.3} & 0.436 & \underline{13194.8} & 0.426 \\
& EvoX          & 19643.8 & 0.440 & 19637.2 & 0.336 \\
& AdaEvolve     & \textbf{9465.8} & 0.388 & 17679.7 & 0.246 \\
\rowcolor{yellow!12}
\cellcolor{white}& \textbf{LabBook (Ours)} & \textbf{9465.8} & 0.180 & \textbf{9465.8} & 0.467 \\
\midrule
\multirow{4}{0.25\textwidth}{Operator Scheduling}
& OpenEvolve    & 354 & 0.200 & \underline{200} & 0.268 \\
& EvoX          & \underline{199} & 0.208 & \underline{200} & 0.388 \\
& AdaEvolve     & \textbf{193} & 0.369 & \underline{200} & 0.335 \\
\rowcolor{yellow!12}
\cellcolor{white}& \textbf{LabBook (Ours)} & \textbf{193} & 0.244 & \textbf{196} & 0.571 \\
\midrule
\multirow{4}{0.25\textwidth}{Pedigree}
& OpenEvolve    & \textbf{5} & 0.377 & 7 & 0.540 \\
& EvoX          & \underline{10} & 0.369 & \underline{6} & 0.541 \\
& AdaEvolve     & \textbf{5} & 0.371 & \underline{6} & 0.537 \\
\rowcolor{yellow!12}
\cellcolor{white}& \textbf{LabBook (Ours)} & \textbf{5} & 0.267 & \textbf{5} & 0.489 \\
\bottomrule
\end{tabular}
\end{table*}

\paragraph{Mathematics.}
With DeepSeek, LabBook achieves both the highest score and the lowest
cost on Circle Packing and Heilbronn Convex. With Qwen, it remains within 1.7\%
and 0.2\% of the best score on the two tasks, respectively, while operating at
the same sub-dollar cost scale as the baselines. Thus, the mathematical results
remain competitive across both backbones.

\paragraph{ADRS.}
On the three systems tasks, LabBook attains the best or tied-best score
with both backbones. With DeepSeek, it simultaneously reduces cost relative to
AdaEvolve and OpenEvolve on EPLB and Transaction Scheduling, and costs less than
AdaEvolve on PRISM. Qwen requires more expensive iterations, but its total
costs remain in the same sub-dollar range as the baselines. Beyond relative
ranking, both backbones attain the proved optimum of
$26.2560$ on PRISM. 

\paragraph{HeuriGym.}
LabBook obtains the best or tied-best result on all four HeuriGym tasks
with both backbones. On DeepSeek, it also has the lowest cost on E-Graph
Extraction and Pedigree and comparable cost on Operator Scheduling and
Pickup--Delivery with Time Windows. On Qwen, its costs remain comparable to the
other methods, while it achieves the lowest unnormalized objective on E-Graph
Extraction, Operator Scheduling, and Pedigree. Relative to HeuriGym's released expert
references, the DeepSeek run matches the aggregate E-Graph, Operator
Scheduling, and Pedigree values exactly.

\subsection{Ablation Studies}

\paragraph{Ablating memory and retrieval.}
We isolate the two components of history construction using DeepSeek V4 Flash
for 30 iterations. \emph{No Retrieval} retains the rewritten LabBook
but removes access to the experimental log, whereas \emph{No Persistent
Memory} retains retrieval but resets the compact state between iterations. We also include three broader alternatives: Best-of-N generates independent
memory-free proposals. \emph{Top-score Retrieval} retains LabBook and the latest experiment but replaces agent-controlled retrieval with the complete program and evaluator feedback of the highest-scoring experiment. \emph{Full History} places every previous
program and evaluation in context.

\begin{figure*}[t]
    \centering
    \includegraphics[width=0.94\textwidth]{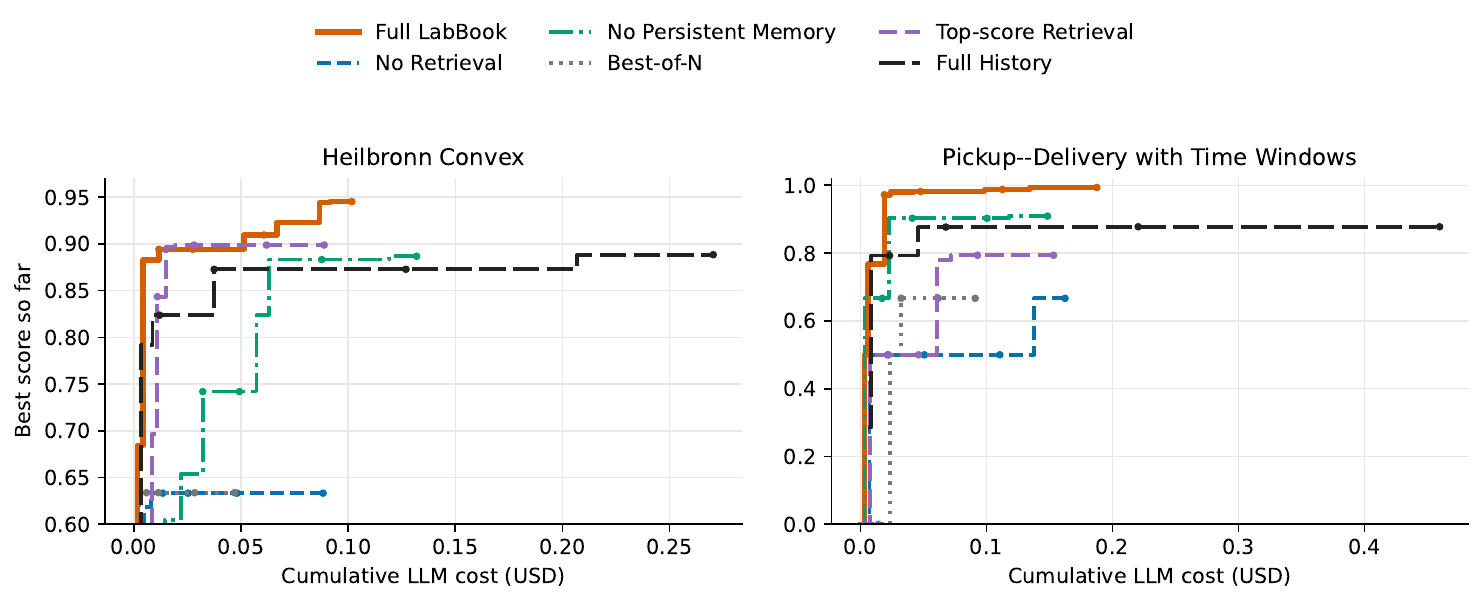}
    \caption{Memory and retrieval ablation over 30 iterations with DeepSeek V4
    Flash. Full LabBook is shown as a solid line; component ablations,
    Best-of-N, Top-score Retrieval, and Full History use distinct dashed styles.
    Markers denote iterations 5, 10, 20, and 30.}
    \label{fig:memory-retrieval-ablation}
\end{figure*}

In Figure~\ref{fig:memory-retrieval-ablation}, on Heilbronn
Convex, Full LabBook continues from $0.8942$ at iteration 5 to $0.9452$
at iteration 30, while No Retrieval and Best-of-N plateau near $0.633$ and No
Persistent Memory, Top-score Retrieval, and Full History reach $0.8868$,
$0.8988$, and $0.8885$, respectively. On Pickup--Delivery with Time Windows,
Full improves from $0.9722$ to $0.9935$, compared with $0.6665$ without
retrieval, $0.9091$ without persistent memory, $0.6667$ for Best-of-N,
$0.7936$ for Top-score Retrieval, and $0.8777$ for Full History.
We also run every
ablation on all nine tasks. Table~\ref{tab:memory-ablation-aggregate}
summarizes the complete evaluation.

Because their native scores have different units, we divide each arm's final score on each task by the best score attained
by any compared arm on that task, and then average over tasks. In particular, E-Graph Extraction, Operator Scheduling, and Pedigree use the normalized HeuriGym score rather than the lower-is-better native objectives. This normalized
mean is used only for cross-task aggregation; the trajectories in
Figure~\ref{fig:memory-retrieval-ablation} and the scores in
Table~\ref{tab:cross-domain-results} retain their task-native units. Full
LabBook achieves the highest normalized mean. Best-of-N is less
expensive but substantially lower in solution quality, while Full History is
more than twice as expensive without improving the aggregate result. Together,
the results show that persistent memory is most effective when it can direct
access to exact historical evidence.

\begin{table}[t]
    \centering
        \caption{Aggregate memory and retrieval ablation over nine tasks at 30
    iterations. Normalization is performed independently per task before
    averaging.}
    \label{tab:memory-ablation-aggregate}
    \small
    \setlength{\tabcolsep}{3.5pt}
    \begin{tabular}{lcccccc}
        \toprule
        & Ours & No Retrieval & No Persistent & Best-of-N & Top-score & Full History \\
        \midrule
        Normalized mean & \textbf{0.9982} & 0.9057 & 0.9789 & 0.8733 & 0.9411 & 0.9488 \\
        Mean cost (USD) & 0.140 & 0.148 & 0.143 & \textbf{0.118} & 0.150 & 0.315 \\
        \bottomrule
    \end{tabular}

\end{table}

\subsection{Case study: long-horizon recovery on Heilbronn Convex}

The task asks for 13 points that maximize
$ R(P)=\nicefrac{\min_{i<j<k} \operatorname{Area}(p_i,p_j,p_k)}
    {\operatorname{Area}(\operatorname{Conv}(P))},
    \label{eq:heilbronn-objective}
$
where the minimum ranges over all $\binom{13}{3}=286$ triangles. This is a
nonsmooth max--min objective: moving a point can change which triangles are
binding, and a local improvement to one small triangle can expose another as
the new minimum. Early in the run, a three-fold symmetric parameterization crashed because its
optimizer supplied nine bounds for a ten-parameter representation.
Our method treated this as an implementation failure rather than evidence
against the geometric idea: it retained the reliable iteration-2 backbone and
the precise dimensional correction. The next proposal repaired the bounds and
improved the best score from $0.882687$ to $0.894220$.

At iteration 33, the agent obtained a score of $0.945243$ using a reliable
multi-stage backbone: soft-min continuation, a low-dimensional polar
construction, exact-objective Nelder--Mead refinement, pattern search, and
simulated annealing. The following 43 attempts explored alternatives but did
not improve the incumbent. Our method nevertheless retained both sides of
this history: iteration 33 as the strongest reusable implementation and the
later failures as evidence against several directions.

\begin{figure}[t]
    \centering
    \includegraphics[width=0.75\linewidth]{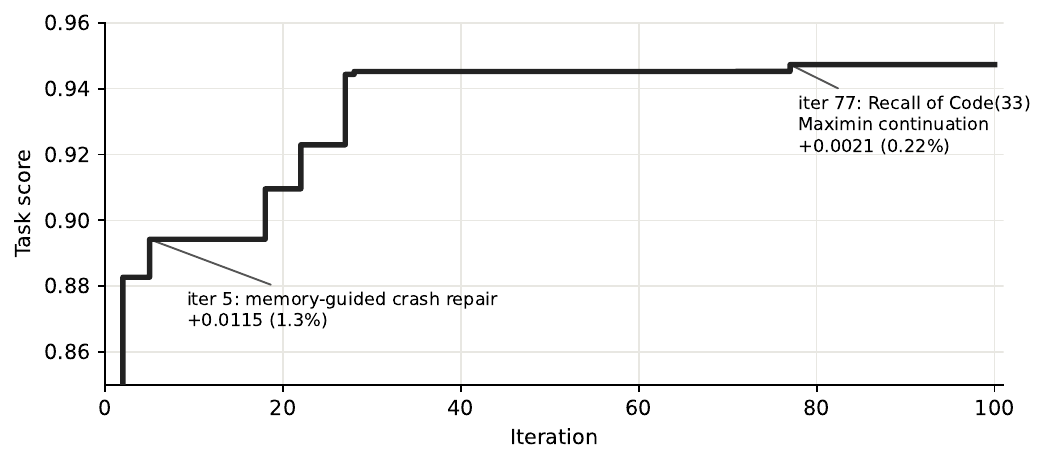}
    \caption{Iteration--performance trajectory for the Heilbronn Convex case
    study. The vertical axis begins at $0.85$ to emphasize improvements beyond
    the initial naive-sampling regime. LabBook records and repairs an
    early implementation failure, preserves iteration 33 as a reliable
    backbone, and retrieves that program 44 iterations later to produce a new
    best at iteration 77.}
    \label{fig:heilbronn-case-study}
\end{figure}

At iteration 77, the agent explicitly retrieved \textsc{Code}(33), reaching
back 44 iterations rather than continuing from the immediately preceding
failure. It preserved the proven optimization backbone and introduced two
targeted updates. First, {maximin feasibility continuation} gradually
raises a target $\tau$ and minimizes
$
    \sum_{i<j<k}\left[\max\!\left(0,
    \tau-\operatorname{Area}(p_i,p_j,p_k)\right)\right]^2,
    \label{eq:heilbronn-feasibility}
$
directly pushing every triangle below the target instead of averaging them
through a smooth surrogate. Second, basin hopping applies random perturbations
followed by exact-objective local refinement, allowing the optimizer to leave
the old binding-triangle basin without discarding the incumbent. The resulting
program reaches a new best score of $0.947325$. This trajectory illustrates the two
roles of LabBook: failed experiments remain available as compressed
negative evidence, while indexed retrieval can recover an exact, much older
implementation when a new idea makes it useful again.
Figure~\ref{fig:heilbronn-case-study} visualizes the resulting
iteration--performance trajectory.

\section{Conclusion}
\label{sec:conclusion}

We studied LLM-driven discovery through the mapping from experimental history
to the context for the next proposal. LabBook realizes this mapping
with a simple separation: a complete, append-only experimental log preserves
exact evidence, while an agent-maintained state summarizes what is currently
known and guides selective retrieval. Across 58 problems, the resulting harness provides a competitive cost--quality trade-off across two language-model backbones. These results suggest that effective discovery need not depend on
increasingly elaborate search structures. Extending this design to noisy, partially observed, and non-programmatic scientific workflows
is a promising direction for future work.






\bibliography{iclr2027_conference}
\bibliographystyle{iclr2027_conference}

\appendix
\section{Appendix}
\label{sec:appendix}

This appendix provides additional details supporting the main paper.
Section~\ref{app:benchmark-composition} documents the benchmark composition and
the selected Frontier-CS problems. Section~\ref{app:additional-metrics}
reports token usage and cost accounting, and Section~\ref{app:memory-size}
measures the growth of the experimental log and LabBook.
Section~\ref{app:frontier-median} gives the median Frontier-CS results.
Finally, Section~\ref{app:prompts} presents the agent prompts and retrieval
protocol, and Section~\ref{app:memory-trajectory} follows the contents of
LabBook across one representative discovery trajectory.

\subsection{Benchmark Composition}
\label{app:benchmark-composition}

Table~\ref{tab:benchmark-composition} summarizes the 58 open-ended problems in
our evaluation. The nine problems from AlphaEvolve, ADRS, and HeuriGym are
listed individually in the table. Our Frontier-CS sample covers all three
problem types adopted by EvoX \citep{liu2026evox}: 10 optimization, 18
constructive, and 21 interactive problems. Table~\ref{tab:frontier-selected-ids}
lists the selected problem identifiers by category.

\begin{table}[t]
\centering
\caption{Algorithmic and research problem benchmarks. We evaluate 58
open-ended problems: two mathematical problems studied by AlphaEvolve, three
systems problems from ADRS, four heuristic-design problems from HeuriGym, and
49 problems sampled from Frontier-CS. Frontier-CS categories follow EvoX
\citep{liu2026evox}.}
\label{tab:benchmark-composition}
\small
\setlength{\tabcolsep}{5pt}
\begin{tabular}{p{0.29\linewidth}cp{0.60\linewidth}}
\toprule
Category & Count & Description \\
\midrule
\multicolumn{3}{l}{\textbf{AlphaEvolve}~\citep{novikov2025alphaevolve}} \\
\texttt{circle\_packing} & 1 &
Place 26 non-overlapping circles in a unit square to maximize the sum of their
radii. \\
\texttt{heilbronn\_convex} & 1 &
Place points in a convex region to maximize the minimum area of any triangle
formed by three points. \\
\midrule
\multicolumn{3}{l}{\textbf{ADRS}~\citep{cheng2025barbarians}} \\
\texttt{eplb} & 1 &
Replicate and place mixture-of-experts experts to balance load across GPUs. \\
\texttt{prism} & 1 &
Place multiple models on GPUs to minimize worst-case memory pressure. \\
\texttt{txn\_scheduling} & 1 &
Order conflicting database transactions to minimize execution makespan. \\
\midrule
\multicolumn{3}{l}{\textbf{HeuriGym}~\citep{chen2026heurigym}} \\
\texttt{egraph\_extraction} & 1 &
Extract a low-cost acyclic expression graph from an e-graph. \\
\texttt{operator\_scheduling} & 1 &
Construct a feasible low-latency schedule under dependency and resource
constraints. \\
\texttt{pedigree} & 1 &
Assign genotypes satisfying Mendelian constraints while minimizing changes. \\
\texttt{pickup\_delivery} & 1 &
Construct capacity- and time-window-feasible vehicle routes with low total
distance. \\
\midrule
\multicolumn{3}{l}{\textbf{Frontier-CS}~\citep{mang2025frontiercs}} \\
Optimization & 10 &
Develop a program that searches over candidate decisions to improve a scalar
evaluation objective while respecting the task's feasibility and resource
limits. \\
Constructive & 18 &
Generate a structured artifact---such as a packing, graph, or
expression---whose components jointly satisfy a set of global constraints. \\
Interactive & 21 &
Design an adaptive program that gathers information about a hidden instance
through successive queries and solves it with as little interaction as
possible. \\
\bottomrule
\end{tabular}

\end{table}

\begin{table}[t]
\centering
\caption{The 49 selected Frontier-CS problems, grouped using the category
definitions adopted by EvoX \citep{liu2026evox}.}
\label{tab:frontier-selected-ids}
\small
\setlength{\tabcolsep}{5pt}
\begin{tabular}{lcp{0.70\linewidth}}
\toprule
Category & Count & Selected Frontier-CS problem IDs \\
\midrule
Optimization & 10 &
2, 16, 42, 44, 48, 50, 58, 305, 311, 312 \\
Constructive & 18 &
0, 3, 5, 13, 60, 72, 73, 83, 85, 89, 178, 179, 180, 187, 210,
217, 227, 239 \\
Interactive & 21 &
14, 101, 107, 111, 117, 122, 138, 144, 151, 153, 154, 155, 159,
161, 162, 165, 168, 169, 170, 247, 249 \\
\bottomrule
\end{tabular}
\end{table}

\subsection{Token Usage and Pricing}
\label{app:additional-metrics}

To understand the source of the observed cost--performance differences, we
compare the token usage and achieved score of the methods at approximately
matched expenditure within each backbone. Table~\ref{tab:token-usage} reports
mean per-problem results on Frontier-CS. For DeepSeek V4 Flash, we use a common
cost ceiling of \$0.35 per problem and select the final displayed 10-iteration
checkpoint below it. This selects iteration 30 for LabBook, iteration 60 for
AdaEvolve, and iteration 50 for EvoX and OpenEvolve. For Qwen 3.6 Flash, we
analogously use a \$1.10 ceiling and select the final available displayed
checkpoint below it: iteration 70 for LabBook and iteration 100 for the
baselines. The thresholds are backbone-specific because the providers use
different rate cards and cache policies; comparisons are made within, rather
than across, backbones. Fresh input excludes tokens reported by the provider as
cache hits, while total tokens sum fresh input, cached input, and output.
Reasoning was disabled and contributed zero tokens in every method.

\begin{table*}[t]
\centering
\caption{Mean per-problem Frontier-CS token usage at the reported checkpoints.
Token counts are rounded to the nearest thousand and cost is in USD.}
\label{tab:token-usage}
\small
\setlength{\tabcolsep}{3.5pt}
\begin{tabular}{llrrrrrrr}
\toprule
Backbone & Method & Iter. & Fresh in & Cached in & Output & Total & Cost & Mean score \\
\midrule
\multirow{4}{*}{DeepSeek V4 Flash}
& OpenEvolve & 50 & 817K & 164K & 292K & 1,273K & \$0.298 & 33.870 \\
& EvoX        & 50 & 879K & 197K & 283K & 1,358K & \$0.302 & 33.988 \\
& AdaEvolve   & 60 & 745K & 476K & 391K & 1,612K & \$0.348 & 37.606 \\
& LabBook     & 30 & 299K & 1,426K & 427K & 2,152K & \$0.306 & \textbf{39.572} \\
\midrule
\multirow{4}{*}{Qwen 3.6 Flash}
& OpenEvolve & 100 & 2,364K & 0 & 555K & 2,919K & \$1.068 & 28.048 \\
& EvoX        & 100 & 2,296K & 0 & 484K & 2,781K & \$0.976 & 24.992 \\
& AdaEvolve   & 100 & 1,631K & 0 & 458K & 2,089K & \$0.821 & 33.055 \\
& LabBook     & 70  & 2,202K & 0 & 519K & 2,721K & \$0.997 & \textbf{39.960} \\
\bottomrule
\end{tabular}
\end{table*}

We apply a fixed DeepSeek off-peak rate card uniformly to every API call,
independent of its timestamp: \$0.15 per million fresh-input tokens, \$0.003
per million cached-input tokens, and \$0.60 per million output tokens.
LabBook processes more total tokens, but 66.3\% are low-priced cache
hits because successive retrieval calls extend the same prompt prefix.
For Qwen, we use the OpenRouter rates of \$0.1875 per million input
tokens and \$1.125 per million output tokens. The Qwen endpoint reported zero
cache-read tokens, so all input was charged at the full
rate. Although the two backbones process token volumes of the same order of
magnitude, Qwen's measured cost is higher primarily because its provider does
not expose cache-read discounts. The
cache difference is especially consequential for LabBook's multi-call
retrieval loop.

\subsection{Experimental-Log Growth and Memory Compression}
\label{app:memory-size}

Table~\ref{tab:memory-size} compares the growth of the append-only experimental
log with the size of the rewritten LabBook. We measure UTF-8 payload
size at iterations 10, 30, and 50. The Frontier-CS row averages over the 49
problems; each remaining row uses the
corresponding DeepSeek run. The experimental log grows continually as complete
records are appended, whereas LabBook does not grow in proportion to
the history. At iteration 50, it remains between 1.38 and 4.89 kB while
representing logs between 460.8 and 1450.5 kB, yielding log-to-memory ratios of
111--594$\times$.

\begin{table*}[t]
\centering
\caption{Growth of the complete experimental log and the compact persistent
memory over 50 iterations. Sizes are decimal kilobytes. The final column is
the ratio between the cumulative log and LabBook at iteration 50.}
\label{tab:memory-size}
\small
\setlength{\tabcolsep}{4pt}
\begin{tabular}{lrrrrrrr}
\toprule
\multirow{2}{*}{Task} &
\multicolumn{3}{c}{Experimental log (kB)} &
\multicolumn{3}{c}{LabBook (kB)} &
\multirow{2}{*}{$|\mathcal{H}_{50}|/|m_{50}|$} \\
\cmidrule(lr){2-4}\cmidrule(lr){5-7}
& Iter. 10 & Iter. 30 & Iter. 50 & Iter. 10 & Iter. 30 & Iter. 50 & \\
\midrule
Frontier-CS (mean)              & 124.9 & 329.4 & 543.3  & 4.20 & 4.84 & 4.89 & 111.2$\times$ \\
Circle Packing                  & 79.6  & 288.3 & 581.4  & 1.34 & 1.75 & 2.26 & 257.8$\times$ \\
Heilbronn Convex                & 67.0  & 257.7 & 460.8  & 0.96 & 2.23 & 1.86 & 247.5$\times$ \\
EPLB                            & 147.7 & 351.7 & 548.1  & 3.11 & 1.46 & 1.38 & 397.7$\times$ \\
PRISM                           & 117.4 & 392.6 & 609.0  & 2.03 & 2.07 & 1.89 & 322.0$\times$ \\
Transaction Scheduling          & 78.5  & 283.5 & 519.0  & 1.95 & 2.16 & 2.25 & 231.2$\times$ \\
E-Graph Extraction              & 85.2  & 338.4 & 661.7  & 2.55 & 1.42 & 1.61 & 412.0$\times$ \\
Operator Scheduling             & 151.4 & 607.4 & 1111.2 & 2.24 & 2.16 & 1.87 & 594.2$\times$ \\
Pedigree                        & 120.2 & 557.7 & 985.5  & 1.97 & 2.91 & 2.43 & 405.9$\times$ \\
Pickup--Delivery with Time Windows
                                & 194.5 & 784.9 & 1450.5 & 1.57 & 1.62 & 2.60 & 558.1$\times$ \\
\bottomrule
\end{tabular}
\end{table*}

\subsection{Median Frontier-CS Results}
\label{app:frontier-median}

Table~\ref{tab:frontier-median} reports median best-so-far performance at the
final checkpoint shown for each method. We first compute the median across the
49 Frontier-CS problems within each run and then average this statistic across
runs. The median results agree with the mean cost--performance curves in
Figure~\ref{fig:frontier-cost-performance}: LabBook obtains the highest
median score with both backbones. The advantage indicates that the mean
result is not driven only by a small number of unusually large improvements.

\begin{table}[t]
\centering
\caption{Median best-so-far Frontier-CS score at the final checkpoint shown
for each method. We compute the median over problems within each run and then
average across runs. Scores use the benchmark's 0--100 scale.}
\label{tab:frontier-median}
\small
\begin{tabular}{llrr}
\toprule
Backbone & Method & Cost (USD) & Median score \\
\midrule
\multirow{4}{*}{DeepSeek V4 Flash}
& LabBook & 0.463 & \textbf{43.065} \\
& AdaEvolve       & 0.348 & 31.792 \\
& EvoX            & 0.370 & 27.054 \\
& OpenEvolve      & 0.365 & 27.708 \\
\midrule
\multirow{4}{*}{Qwen 3.6 Flash}
& LabBook & 1.001 & \textbf{37.000} \\
& AdaEvolve       & 0.821 & 22.715 \\
& EvoX            & 0.976 & 6.956 \\
& OpenEvolve      & 1.068 & 7.900 \\
\bottomrule
\end{tabular}
\end{table}

\subsection{Prompt Templates}
\label{app:prompts}

\lstdefinestyle{discoveryblock}{
    basicstyle=\ttfamily\footnotesize,
    breaklines=true,
    breakatwhitespace=true,
    columns=fullflexible,
    keepspaces=true,
    showstringspaces=false,
    frame=single,
    framerule=0.4pt,
    rulecolor=\color{black!35},
    backgroundcolor=\color{black!2},
    framesep=4pt,
    aboveskip=6pt,
    belowskip=6pt
}

We factor the agent prompt into the three functional blocks that correspond to
Algorithm~\ref{alg:discovery-memory}: the shared discovery policy, the
read-only retrieval protocol, and the joint program--memory output contract.
Text in angle brackets denotes a value supplied by the harness at runtime. The
task statement and latest iteration context are omitted because they
instantiate the variables already defined in Section~\ref{sec:method}.

\paragraph{Shared discovery-agent prompt.}
This block governs the agent decisions in both the retrieval loop and the final
proposal step of Algorithm~\ref{alg:discovery-memory}.

\begin{lstlisting}[style=discoveryblock]
You are the discovery agent in an iterative program-discovery process.
Your next program will be executed and scored exactly as written. Use the latest evaluated program and feedback, LabBook, and the supplied progress signals to decide what experiment to run next. You may inspect exact evidence from earlier iterations through read-only tools. Reason before every action, and do not guess when the relevant evidence can be checked.

1. ESTABLISH THE REFERENCE. Identify the best evidence available so far. If the latest attempt regressed, do not silently discard an earlier useful solution. Inspect its exact code or result when necessary.

2. DIAGNOSE THE LATEST ATTEMPT. Decide whether its limitation is an
implementation error, an insufficiently tested refinement, or an exhausted approach. Use evaluator feedback and prior evidence rather than score alone.

3. CHOOSE THE NEXT MOVE. Refine an approach when a concrete improvement
remains untested. If repeated faithful refinements have plateaued, redirect to a substantively different approach while retaining lessons established
by earlier experiments. Every iteration must propose a concrete new program; never declare an imperfect solution final merely because progress has stalled.

4. UPDATE LabBook. Write the updated long-term memory in plain text. For each approach tried, record its idea, best score, status
(promising, partial, or ruled out), the decisive implementation details,
and the key lesson.
\end{lstlisting}

\paragraph{Read-only retrieval protocol.}
This block specifies the actions and observations in the inner loop of
Algorithm~\ref{alg:discovery-memory}.

\begin{lstlisting}[style=discoveryblock]
To inspect the experimental log, respond with exactly one action:

  THOUGHT: <why this evidence is needed>
  ACTION: history()
  ACTION: code(<iteration>)
  ACTION: result(<iteration>)
  ACTION: done()

history() returns a lightweight index:
  OBSERVATION (history):
    iter <i>: score=<score>

code(i) returns the exact program evaluated at iteration i:
  OBSERVATION (code at iter <i>, score <score>):
    <complete program>

result(i) returns the corresponding evaluator evidence:
  OBSERVATION (result at iter <i>):
    score=<score> verdict=<verdict> approach=<approach>
    judge_feedback: <full evaluator feedback>

done() ends retrieval immediately. Otherwise, each observation is appended
to the active context and the agent may issue another action, up to the
tool-call limit.
\end{lstlisting}

\paragraph{Joint proposal and memory-update contract.}
This block corresponds to the joint update $(x_t,m_t)$ in
Algorithm~\ref{alg:discovery-memory}.

\begin{lstlisting}[style=discoveryblock]
When the available evidence is sufficient, output exactly:

  THOUGHT: <diagnosis and the change made in this attempt>
  FINAL:
  ===CODE===
  ```<language>
  <complete self-contained program, ready to execute>
  ```
  ===MEMORY===
  <the complete rewritten LabBook>

The program must be complete rather than a patch or excerpt. MEMORY contains
the updated long-term memory and must follow the memory requirements above.
\end{lstlisting}

\subsection{Evolution of LabBook}
\label{app:memory-trajectory}

We use Heilbronn Convex to illustrate how LabBook changes over a run.
The following snapshots reproduce the updated memory after three selected checkpoints.

\medskip
\noindent\textbf{$m_5$: retaining a failed but untested idea.}\par
\nopagebreak[4]\smallskip
\begin{lstlisting}[style=discoveryblock,basicstyle=\ttfamily\scriptsize]
- Iter 1: unit-square softmin+L-BFGS-B, weak polish. 0.7.
- Iter 2 (BEST, 0.8827): softmin(log-sum-exp) continuation over
  T=[0.02..0.0005] + L-BFGS-B, 12 random + triangular-grid +
  boundary-heavy + 3-fold starts, then exact-ratio pattern-search
  polish (8 dirs, step 0.05->1e-6), then 40 perturbation restarts.
  min_area_normalized approximately 0.0272.
- Iter 4 (0.8786): SA directly on true ratio; FLAT vs iter 2 -> SA
  on raw min() not better than softmin continuation.
- Iter 5 (0, CRASH): added 3-fold symmetric reduced-dim engine but
  `_sym_optimize` had bounds length 9 while `_sym_expand` needs 10
  params (center 2 + 4 orbits x 2). Pure implementation bug, idea
  untested. FIXED here: bounds = [(-1.0,2.0)]*10.
- Lesson: iter-2 core is the reliable backbone (keep faithfully).
  Symmetric engine now runs correctly; if still flat next time, fix
  hull to an equilateral triangle and optimize 13 points inside it.
- Interface: heilbronn_convex13() -> np.ndarray (13,2).
\end{lstlisting}

\medskip
\noindent\textbf{$m_{20}$: consolidating negative evidence.}\par
\nopagebreak[4]\smallskip
\begin{lstlisting}[style=discoveryblock,basicstyle=\ttfamily\scriptsize]
- BEST 0.909548 (iters 18 & 19, min_area_norm=0.0281386).
  Benchmark=0.030936889 -> ratio 0.9095. Gap approximately 9%.
- DECISIVE CONSTRUCTION (iter 18, KEEP FAITHFULLY): `_areas` via
  cross product; `_hull_area` = ConvexHull.volume; `_ratio =
  min_area/hull_area`; `_softmin_neg =
  -T*(m+log(sum(exp(z-m))))` with z=-areas/T; `_make_starts`
  (12 random + hex4x4 normalized + 6 lattice-perturb sigma=0.04
  + 6 boundary-biased + 6 threefold-folded); temps
  [0.02,0.01,0.005,0.002,0.001,0.0005]; L-BFGS-B box [0,1]^2
  maxiter=400 ftol=1e-15 gtol=1e-12; `_pattern_search` 8 dirs
  step 0.05->1e-6; 40-restart refinement scales
  [0.003,0.008,0.02,0.04]. seed=12345.
- iter 20 (0.8998) RAN but REGRESSED: free-space unboxed engine
  found a worse local optimum and its `fs>s` swap overrode the good
  square result. RULED OUT: free-space/unboxed engine as a
  replacement.
- RULED OUT: iter 17/12 timeouts; iter 16 (0.8749) cached-score bug;
  iter 15 (0) unbounded NM; iter 19 symmetry-orbit engines
  (`c3`, `c6`, `refl`) gave no gain.
- THIS ATTEMPT (iter 21): revert to iter-18 square engine as backbone
  (faithful), add strictly-best-preserving `_extended_polish`
  (120 restarts, scales including 0.001/0.06, temps down to
  0.0002, finer final pattern search step0=0.02). Cannot regress
  below 0.9095 since we always keep max.
- Interface: `heilbronn_convex13() -> np.ndarray (13,2)`,
  deterministic, seed 12345.
- NEXT IF STILL FLAT: Delaunay-circumradius/Voronoi-aware objective;
  explicit 3-fold symmetric polar ansatz; simulated annealing with
  exact-ratio acceptance on a single seed.
\end{lstlisting}

\medskip
\noindent\textbf{$m_{50}$: preserving a distant reusable anchor.}\par
\nopagebreak[4]\smallskip
\begin{lstlisting}[style=discoveryblock,basicstyle=\ttfamily\scriptsize]
- GOAL: maximize min_area_normalized = min triangle area / hull area
  for 13 pts; benchmark 0.030936889. Best that RAN: 0.945 (iter 33).
  Plateau 16+ attempts at 0.945.
- DECISIVE CONSTRUCTION (iter 33, score 0.945) = THE ANCHOR,
  reproduced verbatim this attempt: `_areas` cross-product on
  C(13,3); hull=ConvexHull(P).volume; `_ratio=min(areas)/hull`;
  softmin `_objective` L-BFGS-B bounds [0,1] temps 0.02..0.0005;
  `_make_starts` (8 random + grid + boundary circle + 3-fold 1+4x3
  + 6-fold 1+2x6); `_pattern_search` halving; `_polar_engine` NM
  [(4,3,40,999),(2,6,30,777),(6,2,20,555)];
  `_partial_sym_engine` [(4,3,40,2024),(3,4,30,2025)]; `_anneal`
  exact-ratio SA 40000 iters seed 4242; `_exact_nm_polish` NM
  rounds=3 then 2.
- CRITICAL LESSON: time budget is BINDING. Every attempt (47,48,49,50)
  that ADDED global engines beyond iter 33's set scored 0.857 because
  the extra engines starved the backbone before it reached 0.945.
  Keep engine count/budgets identical to iter 33.
- THIS ATTEMPT (iter 50->51): faithful iter-33 backbone (identical
  budgets) + ONE cheap, structurally different final stage
  `_smooth_slsqp_polish`: gradient-based SLSQP on the smooth softmin
  surrogate (T=0.001,0.0005,0.0002, maxiter=300) to escape the flat
  spots where Nelder-Mead stalls on non-smooth min(). Budget-neutral,
  guarded by strict improvement.
- RULED OUT: basin-hopping global; differential-evolution polar;
  1+6+6 double-ring; exact-3-fold sym3 engine; perturb-then-NM
  multi-start; binding-set NM polish (iter 50, starved backbone ->
  0.857). All either starved budget or failed to beat 0.945.
- NEXT IDEAS IF STILL FLAT: (a) 1+12 single-ring with per-point
  independent angles; (b) coordinate descent in a 3-fold-symmetric
  subspace; (c) reduce backbone restarts to free budget for SA from
  the 0.945 configuration.
\end{lstlisting}

Together, the snapshots show that LabBook acts as a compact decision
state rather than a chronological transcript. At iteration 5 it distinguishes
an implementation failure from evidence against the underlying idea. By
iteration 20, several unsuccessful experiments have been consolidated into
short negative lessons while the strongest construction and its decisive
settings remain explicit. By iteration 50, most low-level chronology has been
dropped, but the reusable iteration-33 anchor, the general time-budget lesson,
and unresolved directions are retained. The changing organization and wording
also show that the state is repeatedly rewritten rather than formed by simply
appending the latest experiment.



\end{document}